\documentclass[10pt,twocolumn,letterpaper]{article}

\usepackage{cvpr}              

\usepackage{graphicx}
\usepackage{amsmath}
\usepackage{amssymb}
\usepackage{booktabs}
\usepackage[normalem]{ulem}

\usepackage[pagebackref,breaklinks,colorlinks]{hyperref}

\usepackage[capitalize]{cleveref}
\crefname{section}{Sec.}{Secs.}
\Crefname{section}{Section}{Sections}
\Crefname{table}{Table}{Tables}
\crefname{table}{Tab.}{Tabs.}

\usepackage[accsupp]{axessibility} 
\usepackage{float}

\def\cvprPaperID{3560} 
\def\confName{CVPR}
\def\confYear{2023}

\newcommand{\redt}[1]{\textcolor{red}{\textbf{#1}}}
\newcommand{\bluet}[1]{\textcolor{blue}{\textbf{#1}}}

\begin{document}

\linespread{0.93}

\title{Multi-modal Learning with Missing Modality via \\ Shared-Specific Feature Modelling}

\author{Hu Wang$^*$, \ Yuanhong Chen$^*$, \ Congbo Ma$^*$, \ Jodie Avery$^*$, \ Louise Hull$^*$, \ Gustavo Carneiro$^\circ$\\
$^*$The University of Adelaide, Adelaide, Australia \\
$^\circ$Centre for Vision, Speech and Signal Processing, University of Surrey, UK\\
{\tt\small $^*$\{firstname.lastname\}@adelaide.edu.au} \tt\small, $^\circ$g.carneiro@surrey.ac.uk
}
\maketitle

\begin{abstract}
The missing modality issue is critical but non-trivial to be solved by multi-modal models. Current methods aiming to handle the missing modality problem in multi-modal tasks, either deal with missing modalities only during evaluation or train separate models to handle specific missing modality settings. In addition, these models are designed for specific tasks, so for example, classification models are not easily adapted to segmentation tasks and vice versa. In this paper, we propose the Shared-Specific Feature Modelling (ShaSpec) method that is considerably simpler and more effective than competing approaches that address the issues above. ShaSpec is designed to take advantage of all available input modalities during training and evaluation by learning shared and specific features to better represent the input data. This is achieved from a strategy that relies on auxiliary tasks based on distribution alignment and domain classification, in addition to a residual feature fusion procedure. Also, the design simplicity of ShaSpec enables its easy adaptation to multiple tasks, such as classification and segmentation. Experiments are conducted on both medical image segmentation and computer vision classification, with results indicating that ShaSpec outperforms competing methods by a large margin. For instance, on BraTS2018, ShaSpec improves the SOTA by more than 3\% for enhancing tumour, 5\% for tumour core and 3\% for whole tumour. The code repository address is \url{https://github.com/billhhh/ShaSpec/}.\footnote{This work received funding from the Australian Government through the Medical Research Futures Fund: Primary Health Care Research Data Infrastructure Grant 2020 and from Endometriosis Australia. G.C. was supported by Australian Research Council through grant FT190100525.}
\end{abstract}

\vspace{-3mm}

\section{Introduction}
\label{sec:intro}

Recently, multi-modal learning has attracted much attention by research and industry communities in both computer vision and medical image analysis. 
Audio, images and short videos are becoming common types of media 
being used for multiple types
model prediction in many different applications, such as sound source localisation~\cite{chen2021localizing}, self-driving vehicles~\cite{wang2019multi} and vision-and-language  applications~\cite{wu2016ask,wang2020soft}. Similarly, in medical domain, combining different modalities to improve diagnosis accuracy has become increasingly important~\cite{dou2020unpaired,wang2022uncertainty}. 
For instance, Magnetic Resonance Imaging (MRI) is a common tool for brain tumour detection, which does not depend only on one type of MRI image, but on multiple modalities (i.e. Flair, T1, T1 contrast-enhanced and T2). 
However, the multi-modal methods above usually require the completeness of all modalities for training and evaluation, limiting their applicability in real-world with missing-modality challenges when subsets of modalities may be missing during training and testing. 

Such challenge has motivated both computer vision~\cite{ma2021smil} and medical image analysis~\cite{havaei2016hemis,dorent2019hetero,shen2019brain,chen2019robust} communities to study missing-modality multi-modal approaches. Wang et al.~\cite{wang2021acn} proposed an adversarial co-training network for missing modality brain tumour segmentation. They specifically introduced a ``dedicated'' training strategy defined by a series of independent models that are specifically trained to each missing situation. 
Another interesting point about all previous methods is that they have been specifically developed either for (computer vision) classification~\cite{ma2021smil} or (medical imaging) segmentation~\cite{havaei2016hemis,dorent2019hetero,shen2019brain,wang2021acn,chen2019robust}, making their extension to multiple tasks challenging.

In this paper, we propose a multi-model learning with missing modality approach, called \underline{\textbf{Sha}}red-\underline{\textbf{Spec}}ific Feature Modelling (ShaSpec), which can handle missing modalities in both training and testing, as well as dedicated training and non-dedicated training\footnote{Non-dedicated training refers to train one model to handle different missing modality combinations.}.
Also, compared with previously models, ShaSpec is designed with a considerably simpler and more effective architecture that explores well-understood auxiliary tasks (e.g., the distribution alignment and domain classification of multi-modal features), which enables ShaSpec to be easily adapted to classification and segmentation tasks. The main contributions are:
\begin{itemize}
\item An extremely simple yet effective multi-modal learning with missing modality method, called Shared-Specific Feature Modelling (ShaSpec), which is based on modelling and fusing shared and specific features to deal with missing modality in training and evaluation and with dedicated and non-dedicated training;
\item To the best of our knowledge, the proposed ShaSpec is the first missing modality multi-modal approach that can be easily adapted to both classification and segmentation tasks given the simplicity of its design.
\end{itemize}

Our results on computer vision classification and medical imaging segmentation benchmarks show that ShaSpec achieves state-of-the-art performance.
Notably, compared with recently proposed competing approaches on BraTS2018, our model shows segmentation accuracy improvements of more than 3\% for enhancing tumour, 5\% for tumour core and 3\% for whole tumour.

\section{Related Work}
\label{sec:related}

\begin{figure*}[htbp]
\centering
\includegraphics[width=0.8\textwidth]{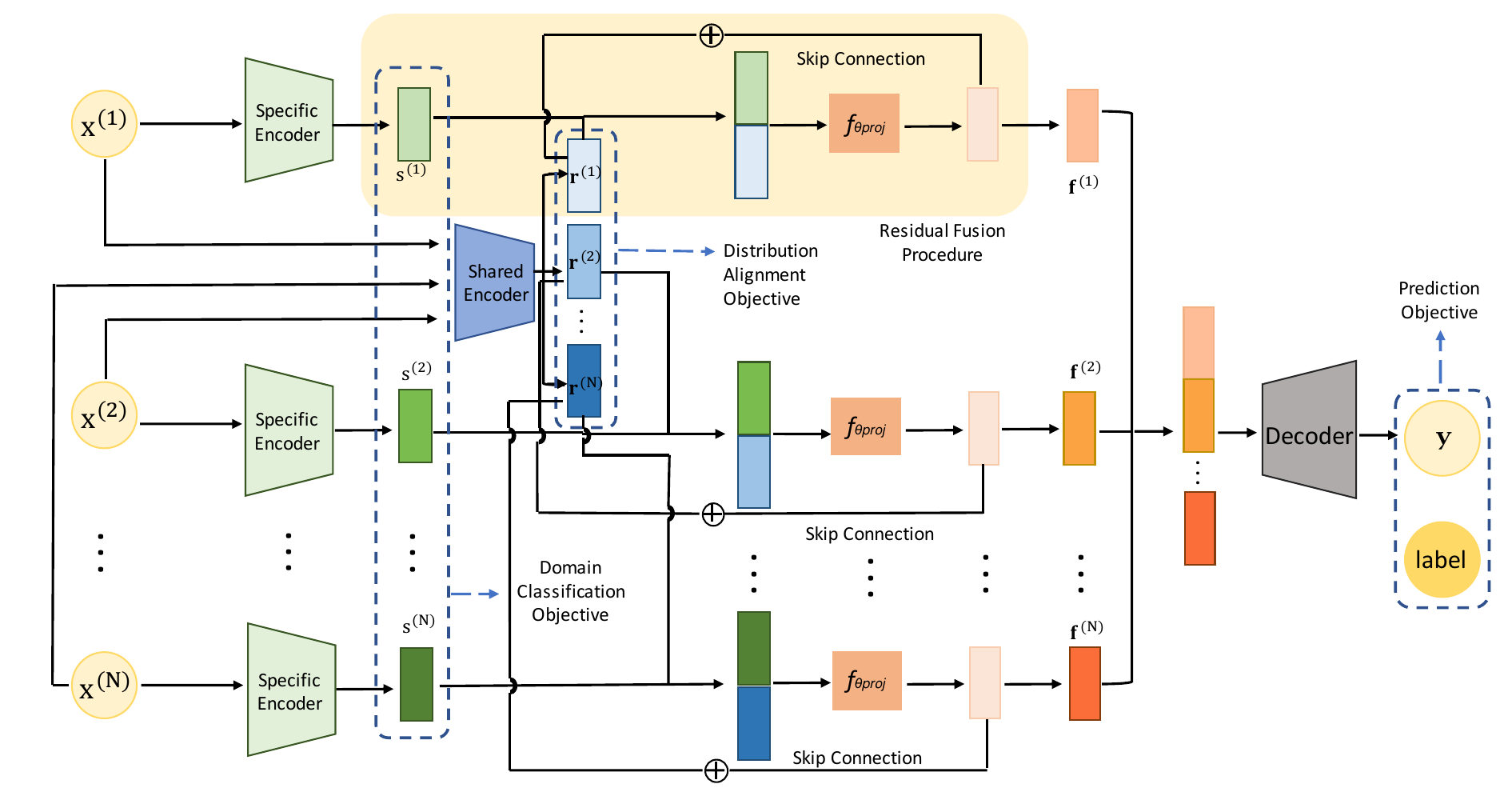}
\vspace{-3mm}
\caption{\textbf{Full-modality training and evaluation of ShaSpec}. All modalities $\{\mathbf{x}^{(i)}\}_{i=1}^{N} \in \mathcal{M}$ are passed through one shared encoder and individual specific encoders to produce the shared features $\{\mathbf{r}^{(i)}\}_{i=1}^{N}$ and specific features $\{\mathbf{s}^{(i)}\}_{i=1}^{N}$, respectively. Then, in a residual learning manner, the shared and specific features are fused with a linear projection $f_{\theta^{\text{proj}}}(\cdot)$ to get the fused features $\{\mathbf{f}^{(i)}\}_{i=1}^{N}$ for decoding. The dashed blue arrows indicate different objective functions.
}
\label{fig:framework_full}
\end{figure*}

\subsection{Multi-modal Learning Models}

Multi-modal learning has attracted increasing attention from the research community. 
In medical image analysis, Dou et al.~\cite{dou2020unpaired} introduced the chilopod-shaped architecture that optimised through modality-dependent feature normalisation and a knowledge distillation objective. 
From a different viewpoint of combining the uncertainty measurement with multi-modal learning, pixel-wise coherency~\cite{monteiro2020stochastic} has been used for multi-modal learning through the optimisation of low-rank covariance metrics. 
Han et al.~\cite{han2021trusted} designed a trusted multi-view classifier by adopting Dirichlet distribution to model the multi-modal uncertainties and fusing features via Dempster's rule. 
Wang et al.~\cite{wang2022uncertainty} introduced an uncertainty-aware multi-modal learning model through cross-modal random network prediction.

In computer vision, Wang et al.~\cite{wang2020deep} combined the idea of channel exchanging and multi-modal learning to fuse features. 
On video/audio classification and retrieval tasks, Patrick et al.~\cite{patrick2020multi, patrick2021space} proposed a self-supervised learning method to train multi-modal models on extra data, which significantly improved model performance. 
Chen et al.~\cite{chen2021localizing} designed a model to improve video-and-sound source localisation accuracy by defining a trusted tri-map middle-ground. 
Jia et al.~\cite{jia2020semi} proposed a model for multi-view learning by constraining the view-specific features to be orthogonal towards view-shared ones. 
Despite providing some improvement over previous approaches, such an orthogonality constraint is fairly strong and not well-motivated by classification/segmentation objectives, so it may hamper the model's ability to learn  semantically rich representations.
The methods above achieve promising results under the completeness assumption of full modalities.
However, in real-world scenarios a subset of modalities may not be available during training and evaluation.

Feature disentanglement methods~\cite{lee2018diverse,liu2022learning} 
aim to model the factors of data variations by learning representations that are modular (each latent dimension denotes a generative factor) and informative (representation has all generative factors) properties. Although our learned shared and specific features are also designed to be modular and informative, we do not aim to perform image (or other input data) reconstruction (i.e., the ShaSpec does not have any generative model) since we only target the learning of optimal representations for the classification and segmentation tasks. This lack of generative requirements significantly simplifies the training of our ShaSpec. 
Also, Jia et al.'s~\cite{jia2020semi} method that learns shared and specific features is based on an arbitrary orthogonality criterion to de-correlate the features, which may not be the optimal way of learning modality-specific and modality-robust features.  
We argue that an optimisation function learns shared features through multi-modal feature distribution alignment and specific features via modality classification will likely lead to better performance. 
An important note about the multi-modal approaches above is that they have been specifically designed for specific tasks, and their adaptation from classification to segmentation, or vice versa, is not straightforward and has not actually been implemented.

\subsection{Addressing Missing Modality in Multi-modal Learning}

To overcome the missing modality issue in multi-modal learning, many methods have been developed.
In computer vision, Ma et al.~\cite{ma2021smil} proposed the SMIL model to deal with missing modality with a meta-learning algorithm that reconstructs the features of the missing modality data. 
Yin et al.~\cite{yin2017unified} aimed to learn a unified subspace for incomplete and unlabelled multi-view data. 
In medical image analysis, Havaei et al.~\cite{havaei2016hemis} developed a model called HeMIS to handle missing modalities by adopting statistical features (mean and variance) for decoding. 
Dorent et al.~\cite{dorent2019hetero}  extended the HeMIS model with a multi-modal variational auto-encoder (MVAE) that produces pixel-wise classifications 
based on mean and variance features. 
Similarly, auto-encoder structures have been adopted to reconstruct the missing modalities in a unsupervised learning scheme~\cite{chartsias2017multimodal,van2018learning}. Multiple approaches~\cite{shen2019brain,wang2021acn,hu2020knowledge} proposed the learning of missing modality features from full modality models to improve the embeddings. 
Zhang et al.~\cite{zhang2022mmformer} introduced a vision transformer architecture for multi-modal brain tumour segmentation that aims to fuse features from all modalities into a set of comprehensive features. 

The aforementioned models are mainly focused on reconstructing the missing modalities/features or introducing sophisticated architectures to solve missing modality problem.
However, all of them neglect an essential point to tackle the missing modality challenge: how to learn the shared (i.e., modality-robust) and specific (i.e., modality-specific) features for optimisation of the model performance.
Chen et al.~\cite{chen2019robust} explored this direction and proposed a feature disentanglement and gated fusion model, called Robust-Mseg, for missing modality multi-modal segmentation. 
However, empirically, the algorithm faces unstable performance when tested on different missing modality scenarios, which we argue is caused by the high complexity of the model. 
Additionally, Robust-Mseg trained the appearance code just for the reconstruction of different modalities, which is weakly linked to the segmentation task.
Moreover, from an implementation perspective\footnote{According to the official released code of \cite{chen2019robust} from \url{https://github.com/cchen-cc/Robust-Mseg}.}, the appearance code generation of the model proposed by~\cite{chen2019robust} contains no missing modality drop and always requires full modality inputs, so it cannot handle missing data in training. 
Similarly to the multi-modal problem, when dealing with missing modality, methods are designed specifically for classification or segmentation and do not generalise to both tasks, which explains the lack of methods that can handle both tasks.

Our ShaSpec addresses the issues listed above with a model of rather simple but effective architecture. It achieves the goal by learning shared and specific features not only via the main task, but also with a distribution alignment and a domain classification tasks. 
The shared and specific features that contain rich information associated with the main task are finally fed into a decoder for prediction, which can be either classification or segmentation.
Also, our model can handle missing modalities in training/testing and dedicated/non-dedicated missing modality training.

\section{Methodology}
\label{sec:model}

\subsection{Overall Architecture}
\label{subsec:arch}

\begin{figure*}[htbp]
\centering
\includegraphics[width=0.8\textwidth]{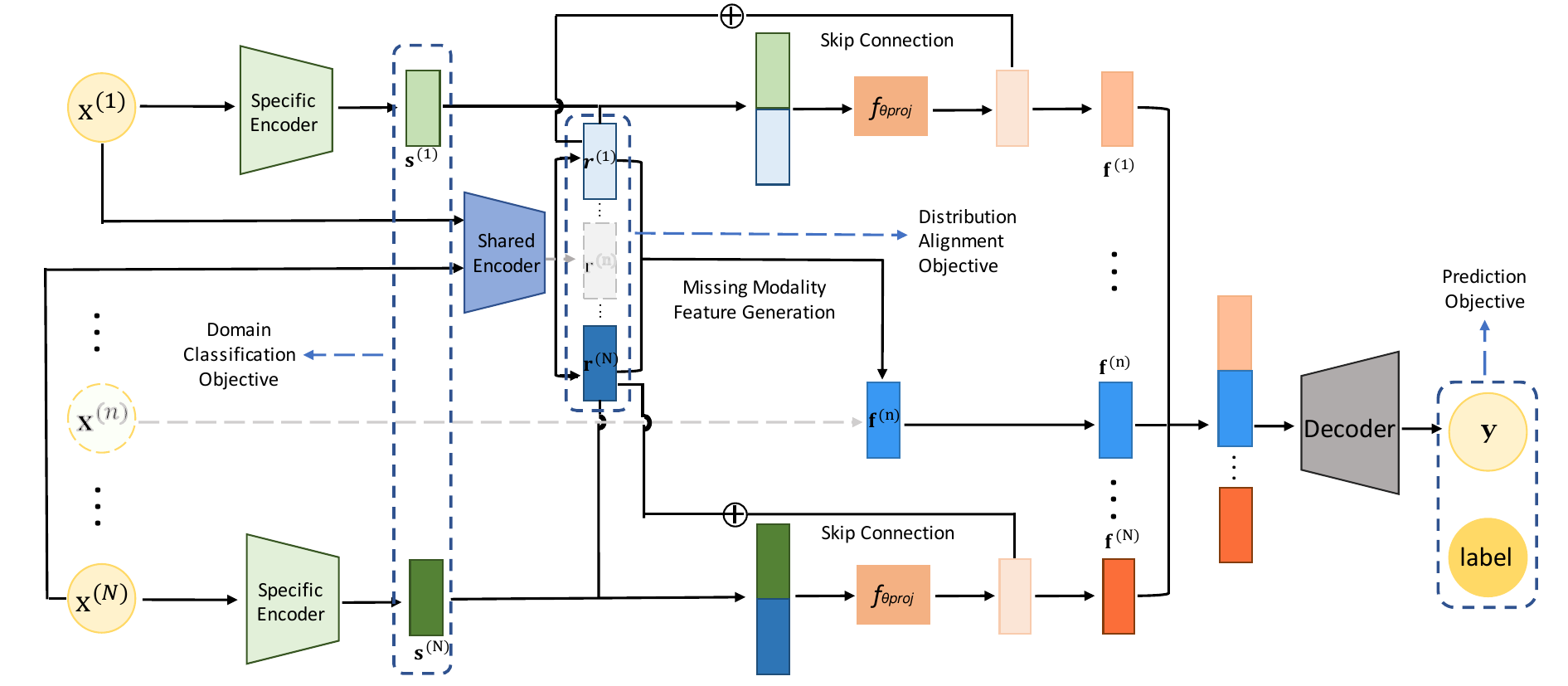}
\vspace{-3mm}
\caption{\textbf{Missing-modality training and evaluation of ShaSpec}. Without losing generality, we assume $\mathbf{x}^{(n)}$ is missing, where $n$ can be $ 1, 2, ..., N$. For available modalities $\mathbf{x}^{(1)}, ..., \mathbf{x}^{(n-1)}, \mathbf{x}^{(n+1)}, ..., \mathbf{x}^{(N)}$, the shared-specific fused features $\mathbf{f}^{(1)}, ..., \mathbf{f}^{(n-1)}, \mathbf{f}^{(n+1)}, ..., \mathbf{f}^{(N)}$ are extracted in the same way as in full modality. But for the missing modality data $\mathbf{x}^{(n)}$, the fused features $\mathbf{f}^{(n)}$ are generated from available shared features $\mathbf{r}^{(1)}, ..., \mathbf{r}^{(n-1)}, \mathbf{r}^{(n+1)}, ..., \mathbf{r}^{(N)}$ via a missing modality feature generation process. The dashed blue arrows indicate different objective functions.
}
\label{fig:framework_missing}
\end{figure*}

Let us represent the $N$-modality data with $\mathcal{M}_j =  \{\mathbf{x}^{(i)}_{j} \}_{i=1}^{N}$, where $\mathbf{x}^{(i)}_{j} \in \mathcal{X}$ denotes the $j^{th}$ data sample and the superscript $^{(i)}$ indexes the modality. To simplify the notation, we omit the subscript $j$ when that information is clear from the context.
The label for each set $\mathcal{M}$ is represented by $\mathbf{y} \in \mathcal{Y}$, where $\mathcal{Y}$ represents the space for segmentation maps or classification categories.
The model consists of a shared encoder denoted by $f_{\theta^{\text{sha}}}:\mathcal{X} \to \mathcal{R}$, specific encoders represented by $f^{(i)}_{\theta^{\text{spec}}}:\mathcal{X} \to \mathcal{S}$ for $i \in \{1,...,N\}$, a feature projection layer $f_{\theta^{\text{proj}}}:\mathcal{R} \times \mathcal{S} \to \mathcal{F}$ 
and a decoder $f_{\theta^{\text{dec}}}:\mathcal{F}^{N} \to \mathcal{Y}$.
As shown in~\cref{fig:framework_full} and~\cref{fig:framework_missing}, the architecture suffers slight modifications while training and evaluating with full modalities or missing modalities. Please note that the ShaSpec model works for segmentation and classification, but the decoder in ~\cref{fig:framework_full} and~\cref{fig:framework_missing} only works for segmentation. For classification, the fused features are fed into fully connected (FC) layers. Below, we explain the evaluation and training processes with full and missing modalities. The methodology works for both dedicated and non-dedicated training.

\subsection{Evaluation with Full and Missing Modalities}

As shown in~\cref{fig:framework_full}, in full modality training/evaluation, the process starts with the shared and specific branches running in parallel, with
\begin{equation}
        \mathbf{r}^{(i)} = f_{\theta^{\text{sha}}}(\mathbf{x}^{(i)}), \text{ and }
        \mathbf{s}^{(i)} = f^{(i)}_{\theta^{\text{spec}}}(\mathbf{x}^{(i)}),
        \label{eq:shared_specific_models}
\end{equation}
for $i \in \{1,...,N\}$. Then, in residual fusion procedure, the shared and specific features are concatenated at the input of the projection layer, whose output is added as a residual to the shared features to form the semantically rich modality embedding, as follows:
\begin{equation}
    \mathbf{f}^{(i)} = f_{\theta^{\text{proj}}}(\mathbf{r}^{(i)},\mathbf{s}^{(i)}) + \mathbf{r}^{(i)}.
    \label{eq:feature_project_model}
\end{equation}
The final decoder then takes all embeddings to produce the output, with
\begin{equation}
    \tilde{\mathbf{y}} = f_{\theta^{\text{dec}}}(\mathbf{f}^{(1)},...,\mathbf{f}^{(N)}),
    \label{eq:decoder}
\end{equation}
where $\tilde{\mathbf{y}} \in \mathcal{Y}$.
The intuition of the model above is that after extracting shared and specific features from each modality, the specific features $\{ \mathbf{s}^{(i)}\}_{i=1}^{N}$ present modality-heterogeneous representations, while the shared features  $\{ \mathbf{r}^{(i)}\}_{i=1}^{N}$ capture the consistent features among modalities.

When a subset of the input modalities is missing, then the model architecture changes, as shown in~\cref{fig:framework_missing}.
Without loss of generality, let us assume that the $n^{th}$ modality  $\mathbf{x}^{(n)} \in \mathcal{M}$ is missing. 
For all other available modalities that are not missing, i.e., $\{ \mathbf{x}^{(i)}\}_{i=1,i \ne n}^{N}$, the process to extract $\{\mathbf{f}^{(i)}\}_{i=1,i \ne n}^{N}$ is the same as with the full modality in~\eqref{eq:shared_specific_models}, but for the missing modality data, 
we directly generate the embedding $\mathbf{f}^{(n)}$ from the other available modalities via a missing modality feature generation process, defined by
\begin{equation}\small
    \mathbf{f}^{(n)} = \frac{1}{N-1}\sum_{i=1,i\ne n}^{N} \mathbf{r}^{(i)}.
    \label{eq:feature_generation}
\end{equation}
The model output is then produced with the decoder in~\cref{eq:decoder}.
When we have more than one, but less than $N$ missing modalities, we simply generate their features from the ones available using~\cref{eq:feature_generation} and adapting the factor $\frac{1}{N-1}$ before summation.

\subsection{Training with Full and Missing Modalities}
\label{sec:obj}

For the model training, besides optimising for the main task (segmentation or classification), we introduce two auxiliary tasks, domain classification and distribution alignment, for the learning of the specific and shared feature representations, respectively.

\subsubsection{Domain Classification Objective}

Inspired by the domain adaptation technique from~\cite{ganin2015unsupervised}, we propose to adopt the domain classification objective (DCO) for the specific feature learning. 
The intuition is that if the specific features from a certain modality can be used to classify its domain (e.g., in brain tumour segmentation the domains can be Flair, T1, T1 contrast-enhanced or T2), then these specific features should contain valuable information that is specific for that modality. 
For domain classification, the cross-entropy (CE) loss is used for all available modalities. Formally, we have:
\begin{equation}
\ell_{\mathit{dco}}(\mathcal{D},\theta^{\text{spec}},\theta^{\text{dco}})=-\sum_{j=1}^{|\mathcal{D}|}\sum_{i=1}^{N} (\mathbf{t}^{(i)})^{\top} \log (f_{\theta^{\text{dco}}}(\mathbf{s}_{j}^{(i)})),
\label{eq:l_dco}
\end{equation}
where $\mathbf{t}^{(i)} \in \{0,1\}^{N}$ is the one-hot modality label with 1 at $i^{th}$ position and 0 elsewhere,
$\mathbf{s}_{j}^{(i)}$ denotes the $i^{th}$ modality specific feature of the $j^{th}$ training sample computed from~\cref{eq:shared_specific_models}
(note that $i \in\{1,...,n-1,n+1,...,N\}$ if modality $n$ is missing),
$f_{\theta^{\text{dco}}}:\mathcal{S} \to \Delta^{N-1}$, with $\Delta^{N-1}$ denoting the probability simplex with $N$ classes, and $\mathcal{D}=\{(\mathcal{M}_{j},\mathbf{y}_{j})\}_{j=1}^{|\mathcal{D}|}$ is the training set.

\subsubsection{Distribution Alignment Objective}
\label{subsubsec:dao}

The distribution alignment objective (DAO) is achieved by attempting to confuse the domain classifier by minimising the CE loss:
\begin{equation}
\ell_{\mathit{dao}}(\mathcal{D},\theta^{\text{sha}},\theta^{\text{dao}})=-\sum_{j=1}^{|\mathcal{D}|}\sum_{i=1}^{N} (\mathbf{u}^{(i)})^{\top} \log (f_{\theta^{\text{dao}}}(\mathbf{r}_{j}^{(i)})),
\label{eq:l_dao}
\end{equation}
where $\mathbf{u}^{(i)} = \frac{1}{N}$ is a uniform distribution for all modalities $i \in\{1,...,N\}$ (note that $i \in\{1,...,n-1,n+1,...,N\}$ if modality $n$ is missing), and 
$f_{\theta^{\text{dao}}}:\mathcal{R} \to \Delta^{N-1}$ is the shared feature modality classifier.
In~\cref{eq:l_dao}, if the classification result can not distinguish the $i^{th}$ modality from the others using the shared feature $\mathbf{r}_j^{(i)}$, then this is a robust shared feature representation.

Another option for this distribution alignment objective is the minimisation of the Kullback–Leibler divergence (KL divergence) between probabilities produced by the shared feature representations. To reduce the computation complexity, we project the feature onto a low dimensional space through a simple linear projection, as follows: 
\begin{equation} \small
\ell_{\mathit{dao}}(\mathcal{D},\theta^{\text{sha}},\theta^{\text{dao}})=\sum_{j=1}^{|\mathcal{D}|}\sum_{i,k=1}^{N}
\text{KL}[\sigma(f_{\theta^{\text{dao}}}(\mathbf{r}_{j}^{(i)})), \sigma(f_{\theta^{\text{dao}}}(\mathbf{r}_{j}^{(k)}))]
\label{eq:l_dao2}
\end{equation}
where the $f_{\theta^{\text{dao}}}(\cdot)$ is the linear projection that produces an input for the softmax function $\sigma(\cdot)$, 
$KL(\cdot)$ is the Kullback-Leibler divergence operator.
One more option for the DAO is the pairwise feature similarity, using 
\begin{equation}
\ell_{\mathit{dao}}(\mathcal{D},\theta^{\text{sha}})=\sum_{j=1}^{|\mathcal{D}|}\sum_{i,k=1}^{N}
\|\mathbf{r}_{j}^{(i)} - \mathbf{r}_{j}^{(k)}\|_p,
\end{equation}
where $||\cdot||_p$ denotes the p-norm operator.
In the ablation studies, we test different distribution alignment objectives.

\subsubsection{Overall Objective}

Besides the aforementioned DCO and DAO objectives, the objective for the main task is denoted by $\ell_{\mathit{task}}(.)$ (e.g., cross-entropy loss for classification or Dice loss for segmentation). The overall objective to be minimised is:
\begin{equation}\small
\begin{split}
\ell_{\mathit{tot}}(\mathcal{D},\Theta) = & \ell_{\mathit{task}}(\mathcal{D},\theta^{\text{sha}}, \theta^{\text{spec}},\theta^{\text{proj}},\theta^{\text{dec}}) + \\ & \alpha \ell_{\mathit{dao}}(\mathcal{D},\theta^{\text{sha}},\theta^{\text{dao}}) + \beta \ell_{\mathit{dco}}(\mathcal{D},\theta^{\text{spec}},\theta^{\text{dco}}),
\end{split}
\label{eq:main_loss}
\end{equation}
where $\Theta=\{\theta^{\text{sha}},\theta^{\text{spec}},\theta^{\text{proj}},\theta^{\text{dao}},\theta^{\text{dco}},\theta^{\text{dec}}\}$; $\alpha$ and $\beta$ are trade-off factors between different objective functions. In the ablation studies we test multiple values for $\alpha$ and $\beta$.

When dealing with missing modality $n$ in training, the calculation of the missing modality feature $\mathbf{f}^{(n)}$ follows~\cref{eq:feature_generation}. This allows the optimisation of $\ell_{\mathit{task}}(.)$. For the optimisation of $\ell_{\mathit{dao}}(.)$ and $\ell_{\mathit{dco}}(.)$, the losses from missing modality features $\mathbf{r}_{j}^{(n)}$ and $\mathbf{s}_{j}^{(n)}$ are omitted. Therefore, our proposed framework can seamlessly deal with missing modality issues in both training and evaluation.

\section{Experiments}
\label{sec:exp}

\begin{table*}[htb]\small
\setlength\tabcolsep{1pt}
\begin{center}
\resizebox{1.0\linewidth}{!}{
\begin{tabular}{|cccc|cccc|cc|cccc|cc|cccc|cc|}
\hline
\multicolumn{4}{|c|}{Modalities}               & \multicolumn{6}{c|}{Enhancing tumour}                     & \multicolumn{6}{c|}{tumour Core}                          & \multicolumn{6}{c|}{Whole tumour}                         \\ \hline
Fl         & T1        & T1c      & T2        & UHeMIS & UHVED & RbSeg & mmFm & ShaSpec  & ShaSpec* & UHeMIS & UHVED & RbSeg & mmFm & ShaSpec  & ShaSpec* & UHeMIS & UHVED & RbSeg & mmFm & ShaSpec  & ShaSpec* \\ \hline
$\bullet$ & $\circ$   & $\circ$   & $\circ$   & 11.78   & 23.80   & 25.69     & 39.33    & \bluet{43.52} & \redt{45.11} & 26.06   & 57.90   & 53.57     & 61.21    & \bluet{69.44} & \redt{69.57} & 52.48   & 84.39  & 85.69     & 86.10     & \bluet{88.68} & \redt{88.83} \\
$\circ$   & $\bullet$ & $\circ$   & $\circ$   & 10.16   & 8.60    & 17.29     & 32.53    & \bluet{41.00}    & \redt{42.58} & 37.39   & 33.90   & 47.90      & 56.55    & \bluet{63.18} & \redt{64.53} & 57.62   & 49.51  & 70.11     & 67.52    & \bluet{73.44} & \redt{74.82} \\
$\circ$   & $\circ$   & $\bullet$ & $\circ$   & 62.02   & 57.64  & 67.07     & 72.60     & \bluet{73.29} & \redt{75.80}  & 65.29   & 59.59  & 76.83     & 75.41    & \bluet{78.65} & \redt{81.40}  & 61.53   & 53.62  & 73.31     & 72.22    & \bluet{73.82} & \redt{74.95} \\
$\circ$   & $\circ$   & $\circ$   & $\bullet$ & 25.63   & 22.82  & 28.97     & 43.05    & \redt{46.31} & \bluet{46.21} & 57.20    & 54.67  & 57.49     & 64.20     & \bluet{69.03} & \redt{69.05} & 80.96   & 79.83  & 82.24     & 81.15    & \bluet{83.99} & \redt{84.90}  \\
$\bullet$ & $\bullet$ & $\circ$   & $\circ$   & 10.71   & 27.96  & 32.13     & 42.96    & \bluet{44.76} & \redt{44.81} & 41.12   & 61.14  & 60.68     & 65.91    & \bluet{72.67} & \redt{72.77} & 64.62   & 85.71  & 88.24     & 87.06    & \bluet{89.76} & \redt{89.86} \\
$\bullet$ & $\circ$   & $\bullet$ & $\circ$   & 66.10   & 68.36  & 70.30     & 75.07    & \bluet{75.60} & \redt{77.76} & 71.49   & 75.07  & 80.62     & 77.88    & \bluet{84.50} & \redt{84.75} & 68.99   & 85.93  & 88.51     & 87.30     & \bluet{90.06} & \redt{90.12} \\
$\bullet$ & $\circ$   & $\circ$   & $\bullet$ & 30.22   & 32.31  & 33.84     & \redt{47.52}    & 47.20 & \bluet{47.22} & 57.68   & 62.70  & 61.16     & 69.75    & \redt{72.93} & \redt{72.93} & 82.95   & 87.58  & 88.28     & 87.59    & \bluet{90.02} & \redt{90.09} \\
$\circ$   & $\bullet$ & $\bullet$ & $\circ$   & 66.22   & 61.11  & 69.06     & 74.04    & \bluet{75.76} & \redt{78.26} & 72.46   & 67.55  & 78.72     & 78.59    & \bluet{82.10} & \redt{82.64} & 68.47   & 64.22  & 77.18     & 74.42    & \bluet{78.74} & \redt{78.88} \\
$\circ$   & $\bullet$ & $\circ$   & $\bullet$ & 32.39   & 24.29  & 32.01     & 44.99    & \bluet{46.84} & \redt{49.87} & 60.92   & 56.26  & 62.19     & 69.42    & \bluet{71.38} & \redt{71.39} & 82.41   & 81.56  & 84.78     & 82.20    & \bluet{86.03} & \redt{86.09} \\
$\circ$   & $\circ$   & $\bullet$ & $\bullet$ & 67.83   & 67.83  & 69.71     & 74.51    & \bluet{75.95} & \redt{78.59} & 76.64   & 73.92  & 80.20     & 78.61    & \bluet{83.82} & \redt{84.08} & 82.48   & 81.32  & 85.19     & 82.99    & \bluet{85.42} & \redt{86.43} \\
$\bullet$ & $\bullet$ & $\bullet$ & $\circ$   & 68.54   & 68.60  & 70.78     & 75.47    & \bluet{76.42} & \redt{78.51} & 76.01   & 77.05  & 81.06     & 79.80    & \bluet{85.23} & \redt{85.36} & 72.31   & 86.72  & 88.73     & 87.33    & \bluet{90.29} & \redt{90.36} \\
$\bullet$ & $\bullet$ & $\circ$   & $\bullet$ & 31.07   & 32.34  & 36.41     & \redt{47.70}    & 46.55 & \bluet{46.56} & 60.32   & 63.14  & 64.38     & 71.52    & \bluet{73.97} & \redt{73.99} & 83.43   & 88.07  & 88.81     & 87.75    & \bluet{90.36} & \redt{90.37} \\
$\bullet$ & $\circ$   & $\bullet$ & $\bullet$ & 68.72   & 68.93  & 70.88     & 75.67    & \bluet{75.99} & \redt{78.15} & 77.53   & 76.75  & 80.72     & 79.55    & \bluet{85.26} & \redt{85.67} & 83.85   & 88.09  & 89.27     & 88.14    & \bluet{90.78} & \redt{90.79} \\
$\circ$   & $\bullet$ & $\bullet$ & $\bullet$ & 69.92   & 67.75  & 70.10     & 74.75    & \bluet{76.37} & \redt{78.35} & 78.96   & 75.28  & 80.33     & 80.39    & \bluet{84.18} & \redt{84.27} & 83.94   & 82.32  & 86.01     & 82.71    & \bluet{86.47} & \redt{86.51} \\
$\bullet$ & $\bullet$ & $\bullet$ & $\bullet$ & 70.24   & 69.03  & 71.13     & 77.61    & \bluet{78.08} & \redt{78.47} & 79.48   & 77.71  & 80.86     & \redt{85.78}    & 85.45 & \bluet{85.75} & 84.74   & 88.46  & 89.45     & 89.64    & \redt{90.88} & \redt{90.88} \\ \hline
\multicolumn{4}{|c|}{Average}                   & 46.10   & 46.76  & 51.02     & 59.85    & \bluet{61.58} & \redt{63.08} & 62.57   & 64.84  & 69.78     & 72.97    & \bluet{77.45} & \redt{77.88} & 74.05   & 79.16  & 84.39     & 82.94    & \bluet{85.92} & \redt{86.26} \\ \hline
\end{tabular}
}\end{center}
\vspace{-3mm}
\caption{Model performance comparison of \textbf{segmentation} Dice score (normalised to 100\%) on BraTS2018 of \textbf{non-dedicated training}. 
ShaSpec and ShaSpec* are the proposed models, with ShaSpec* being the model with prediction smoothness enhancement. The best and second best results for each column within a certain type of tumour are in \redt{red} and \bluet{blue}, respectively.}
\label{tab:brats2018}
\end{table*}

\begin{table*}[htb]\small
\begin{center}
\resizebox{1.0\linewidth}{!}{
\begin{tabular}{|cccc|cc|cc|cc|cc|cc|cc|}
\hline
\multicolumn{4}{|c|}{Modalities}                & \multicolumn{4}{c|}{Enhancing tumour} & \multicolumn{4}{c|}{tumour Core} & \multicolumn{4}{c|}{Whole tumour} \\ \hline
Fl         & T1        & T1c      & T2        & KD-Net   & ACN    & ShaSpec   & ShaSpec*  & KD-Net & ACN   & ShaSpec  & ShaSpec* & KD-Net  & ACN   & ShaSpec  & ShaSpec* \\ \hline
$\bullet$ & $\circ$   & $\circ$   & $\circ$   & 40.99    & 42.77  & \bluet{43.94}  & \redt{43.97}  & 65.97  & 67.72 & \bluet{70.97} & \redt{70.99} & 85.14   & 87.30  & \bluet{89.28} & \redt{89.38} \\
$\circ$   & $\bullet$ & $\circ$   & $\circ$   & 39.87    & 41.52  & \bluet{45.24}  & \redt{46.76}  & 70.02  & \redt{71.18} & 70.28 & \bluet{70.64} & 77.28   & 79.34 & \bluet{79.40}  & \redt{79.50}  \\
$\circ$   & $\circ$   & $\bullet$ & $\circ$   & 75.32    & \bluet{78.07}  & 75.91  & \redt{78.40}   & 81.89  & 84.18 & \bluet{84.19} & \redt{85.47} & 76.79   & \bluet{80.52} & 80.43 & \redt{80.55} \\
$\circ$   & $\circ$   & $\circ$   & $\bullet$ & 39.04    & 42.98  & \bluet{44.54}  & \redt{46.07}  & 66.01  & 67.94 & \redt{70.30}  & \bluet{70.11} & 82.32   & 85.55 & \bluet{85.58} & \redt{85.62} \\ \hline
\multicolumn{4}{|c|}{Average}                   & 48.81    & 51.34  & \bluet{52.41}  & \redt{53.80}  & 70.97  & 72.76 & \bluet{73.92} & \redt{74.30}  & 80.38   & 83.18 & \bluet{83.67} & \redt{83.76} \\ \hline
\end{tabular}
}\end{center}
\vspace{-3mm}
\caption{Model performance comparison of \textbf{segmentation} Dice score (normalised to 100\%) on BraTS2018 of \textbf{dedicated training}.
}
\label{tab:brats2018-dedicated}
\end{table*}

\subsection{Datasets}

We test our multi-modal learning with missing modality method on two datasets, the BraTS2018 for medical image segmentation and Audiovision-MNIST for computer vision classification.
The \textbf{BraTS2018 Segmentation Challenge dataset}~\cite{menze2014multimodal,bakas2018identifying} is used as a multi-modal learning with missing modality brain tumour sub-region segmentation benchmark, where the sub-regions are enhancing tumour (ET), tumour core (TC), and whole tumour (WT).  
BraTS2018 contains 3D multi-modal brain MRIs, including Flair, T1, T1 contrast-enhanced (T1c) and T2 with experienced imaging experts annotated ground-truth. It includes 285 cases for training (210 gliomas with high grade and 75 gliomas with low grade) and 66 cases for evaluation. The ground-truth of training set is publicly available, but the annotations of validation set is hidden and online evaluation\footnote{Online evaluation at \url{https://ipp.cbica.upenn.edu/}.} is required.

\begin{table*}[htb]
\begin{center}
\resizebox{0.7\linewidth}{!}{
\begin{tabular}{|c|cc|cccc|c|}
\hline
Audio rate & LowerB & UpperB & AutoEncoder & GAN & Full2miss & SMIL & ShaSpec \\ \hline
5\%                             & 92.35                        & 98.22                           & 89.78                  & 89.11                   & 90.00                         & 92.89                    & \textbf{93.33}                       \\
10\%                            & 92.35                        & 98.22                           & 89.33                  & 89.78                   & 91.11                         & 93.11                    & \textbf{93.56}                       \\
15\%                            & 92.35                        & 98.22                           & 89.78                  & 88.67                   & 92.23                         & 93.33                    & \textbf{93.78}                       \\
20\%                            & 92.35                        & 98.22                           & 88.89                  & 89.56                   & 92.67                         & 94.44                    & \textbf{94.67} \\ \hline
\end{tabular}
}\end{center}
\vspace{-3mm}
\caption{Model performance comparison of classification accuracy of missing modality (by setting different available audio rates) on Audiovision-MNIST dataset. The lower bound (LowerB) is a LeNet~\cite{lecun1998gradient} network trained with single modality (images only). The upper bound (UpperB) is a model trained with all data modalities (all images and audios). 
The best results for each row are bolded.}
\label{tab:avmnist}
\end{table*}

Our missing modality experiments for computer vision classification are conducted on \textbf{Audiovision-MNIST dataset}~\cite{vielzeuf2018centralnet}. Audiovision-MNIST is a multi-modal dataset consisting of 1500 samples of audio and image files. The images for digits 0 to 9 are of a size
28$\times$28  that come from the MNIST dataset~\cite{lecun1998gradient}. 
The audio dataset contains 1500 audio files that have been collected from the Free Spoken Digits Dataset\footnote{The data information can be found at \url{https://github.com/Jakobovski/free-spoken-digit-dataset}.}. For the representation of audio modality, mel-frequency cepstral coefficients (MFCCs) are adopted to transform each audio sample into size  20$\times$20$\times$1. Following~\cite{ma2021smil}, we split the dataset into 70\% for training and 30\% for evaluation, according to the officially released code\footnote{The code address is \url{https://github.com/mengmenm/SMIL}.}.

\subsection{Implementation Details}

The ShaSpec model has a straightforward training process without much hyperparameter tuning. Implementation details on both datasets are described below.

\noindent\textbf{BraTS2018:} We adopted the 3D UNet (with 3D convolution and normalisation) as our backbone network, where the fusion of shared and specific features happens at the bottom of the UNet structure. 
A stochastic gradient descent optimizer with Nesterov momentum~\cite{botev2017nesterov} of 0.99 is adopted for optimisation. The learning rate is set to $10^{-2}$ at the beginning and decreased with cosine annealing strategy~\cite{loshchilov2016sgdr}. During the non-dedicated training of models, modalities are randomly dropped to simulate the modality-missing situations. 
For dedicated training of models, the missing modalities used for training are the same missing modalities in the evaluation. 
The ShaSpec model is trained for 180,000 iterations using all training data without model selection. We choose L1 loss as our distribution alignment objective and set $\alpha=0.1$, $\beta=0.02$ in~\cref{eq:main_loss}.
Then, we perform the official online evaluation using the segmentation masks produced by ShaSpec.
We run ShaSpec with and without prediction smoothness enhancement to improve segmentation results, when we run our model with this enhancement. We labelled it as ``ShaSpec*'. 
This enhancement connects components within the surrounding voxels with two hops 
being considered as neighbours. 
Moreover, small regions with fewer voxels than a certain threshold are eliminated. By doing so, scattered small regions are cancelled, which in general leads to segmentation improvements.

\noindent\textbf{Audiovision-MNIST:} For the model training on Audiovision-MNIST dataset, we follow the SMIL paper~\cite{ma2021smil} by dropping the sound modality data for a certain percentage and training all models for 60 epochs to keep a fair comparison. 
we adopted the image and sound encoders from SMIL, consisting of networks with a sequence of convolutional layers and fully connected (FC) layers with batch norm and dropout. For the rest of the ShaSpec architecture, after fusing the two modality features, 2 FC layers with dropout are adopted for classification. 
The fusion of shared and specific features happens at the layer before the FC layers. Adam optimizer with $10^{-2}$ weight decay is used for model training. The initial learning rate is set to $10^{-3}$ which is decreased by 10\% every 20 epochs. 

The evaluation of model performance relies on the Dice score for BraTS2018 and classification accuracy for Audiovision-MNIST, where training and evaluation are performed on one 3090Ti NVIDIA Graphics Card.

\subsection{Segmentation Results}

The experimental results for the non-dedicated training (training the model once and evaluating it on different combinations of missing modalities) on BraTS2018 are shown in~\cref{tab:brats2018}, which compares ShaSpec with current state-of-the-art (SOTA) methods, including U-HeMIS~\cite{havaei2016hemis}, U-HVED~\cite{dorent2019hetero}, Robust-MSeg (RbSeg)~\cite{chen2019robust} and mmFormer (mmFm)~\cite{zhang2022mmformer}.
Our proposed ShaSpec shows the best and second best results (the best in 45 out of 48 results) across almost all different combinations and tumour types, as shown in red and blue of~\cref{tab:brats2018}. 
Moreover, the proposed ShaSpec outperforms the competing models by a large margin. For instance, when only T1 is available, ShaSpec surpasses the second best model (mmFormer) by 8.47\% on enhancing tumour, 6.63\% on tumour core and 5.92\% on whole tumour. 
Similarly, when only T1, T1c and T2 are available, we observe improvements of 1.62\% on enhancing tumour, 3.79\% on tumour core and 3.76\% on whole tumour. 
The Prediction Smoothness Enhancement further boosts ShaSpec's performance. 
On average, when compared with the second best competing methods, our model gets 3.23\% performance gain for enhancing tumour, 4.91\% for tumour core and 3.32\% for whole tumour.
Also, in~\cref{tab:brats2018} note that for the enhancing tumour segmentation, the prediction smoothness enhancement is influential, improving from 43.52\% to 45.11\% when only Flair is available, and from 75.76\% to 78.26\% when T1 and T1c are available.
This may be caused by scattered segmentation masks predicted for enhancing tumour. 
Also, T1c contributes more than other modalities for enhancing tumour. For instance, when adding T1c as an available modality, the performance of the model increases greatly. For ShaSpec, we have 73.29\% with T1c only vs. 43.52\% with Flair only. Such observation resonates with the knowledge that enhancing tumour is clearly visible in T1c, but edema is not visible~\cite{chen2019robust}.
Similar results are shown in~\cref{tab:brats2018-dedicated}, where ShaSpec outperforms KD-Net~\cite{hu2020knowledge} and ACN~\cite{wang2021acn} for most of cases (we are best in 11 out of 12 cases). On average, our model surpasses the second best competing method by 2.46\%  on enhancing tumour, 1.54\% on tumour core and 0.58\% on whole tumour.

\subsection{Classification Results}

Following the SMIL setup~\cite{ma2021smil}, we train ShaSpec on both partial and full modality sub-datasets (images and audios). More specifically, full missing modality sub-dataset is formed by setting the audio modality rates in $\{ 5\%, 10\%, 15\%, 20\% \}$, which defines the proportion of the available audio data used for training. For this setup, the visual modality data is fully available. 
In the evaluation phase, only images are fed into the models. 
We compare our model with Auto-encoder~\cite{baldi2012autoencoders}, a model based on Generative Adversarial Networks~\cite{goodfellow2020generative}, a method that distils multi-modality knowledge to train missing modality model~\cite{shen2019brain}, and SMIL~\cite{ma2021smil}. A  LeNet~\cite{lecun1998gradient} network with single modality (images only) serves as the lower bound, and following SMIL, a model trained with full modalities (all images and audios) works as the upper bound.
As shown in~\cref{tab:avmnist}, ShaSpec performs well, especially under extreme missing modality (i.e., small audio rates), where our model achieves 93.33\% accuracy compared with second best model with 92.89\% on audio rate $5\%$. 
With the increasing of the audio rate, the performance of all models improved, with ShaSpec still outperforming all other models. 
We argue that this is due to the outstanding ability of ShaSpec to extract information-rich shared-specific representations from all available modalities and the use of the shared representation to make up for the  missing modalities.

\begin{figure}[htbp]
\begin{center}
\includegraphics[width=0.45\textwidth]{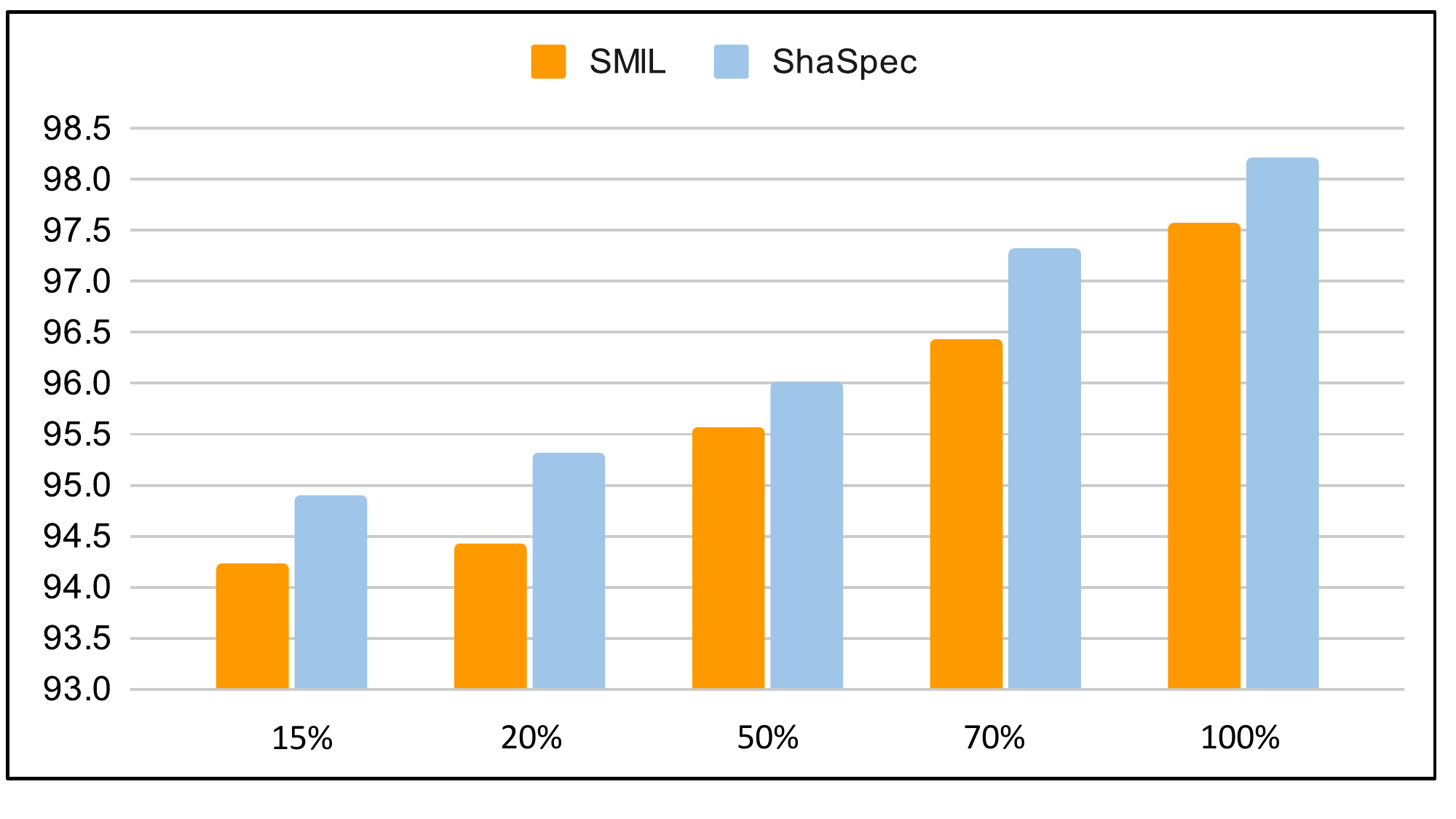}
\end{center}
\vspace{-3mm}
\caption{Model performance with varying rates of missing modality data (for both image and audio) on Audiovision-MNIST.}
\label{fig:avmnist-multimodal}
\end{figure}

To examine the missing modality representation learning of the proposed ShaSpec, we train models for varying rates of both image and audio (rates in 15\%, 20\%, 50\%, 70\% and 100\%) and test with both modalities. The ShaSpec model is compared with SMIL in~\cref{fig:avmnist-multimodal}. For all settings, our model is always superior as shown by the bar chart.

\subsection{Analyses}

\noindent\textbf{The selection of DAO loss function:} As discussed in~\cref{subsubsec:dao}, DAO can rely on a CE loss, KL divergence loss or p-norm distance loss between prediction and ground-truth. The comparisons on BraTS2018 is presented in~\cref{tab:selection-dao} for non-dedicated training, where only T1 is available for evaluation. It shows that the different choices of DAO perform similarly, but the L1 loss has the best results, followed by the KL divergence loss, while CE loss and MSE do not perform as well as the other losses.

\begin{table}[t]\small
\begin{center}
\resizebox{0.9\linewidth}{!}{
\begin{tabular}{|l|cccc|}
\hline
DAO Type        & CE & KL & L1 & MSE \\ \hline
Enhancing Tumour & 41.23                   & 40.41                  & \bluet{42.58}                  & \redt{43.19}                   \\
Tumour Core      & 62.72                   & \bluet{62.83}                  & \redt{64.53}                  & 62.44                   \\
Whole Tumour     & 73.91                   & \bluet{74.19}                  & \redt{74.82}                  & 73.25                 \\ \hline
\end{tabular}
}\end{center}
\vspace{-3mm}
\caption{
Model ablation of different distribution alignment objectives for non-dedicated training, where only T1 is available for testing on BraTS2018. 
}
\label{tab:selection-dao}
\end{table}

\begin{figure}[t]
\begin{center}
\includegraphics[width=0.5\textwidth]{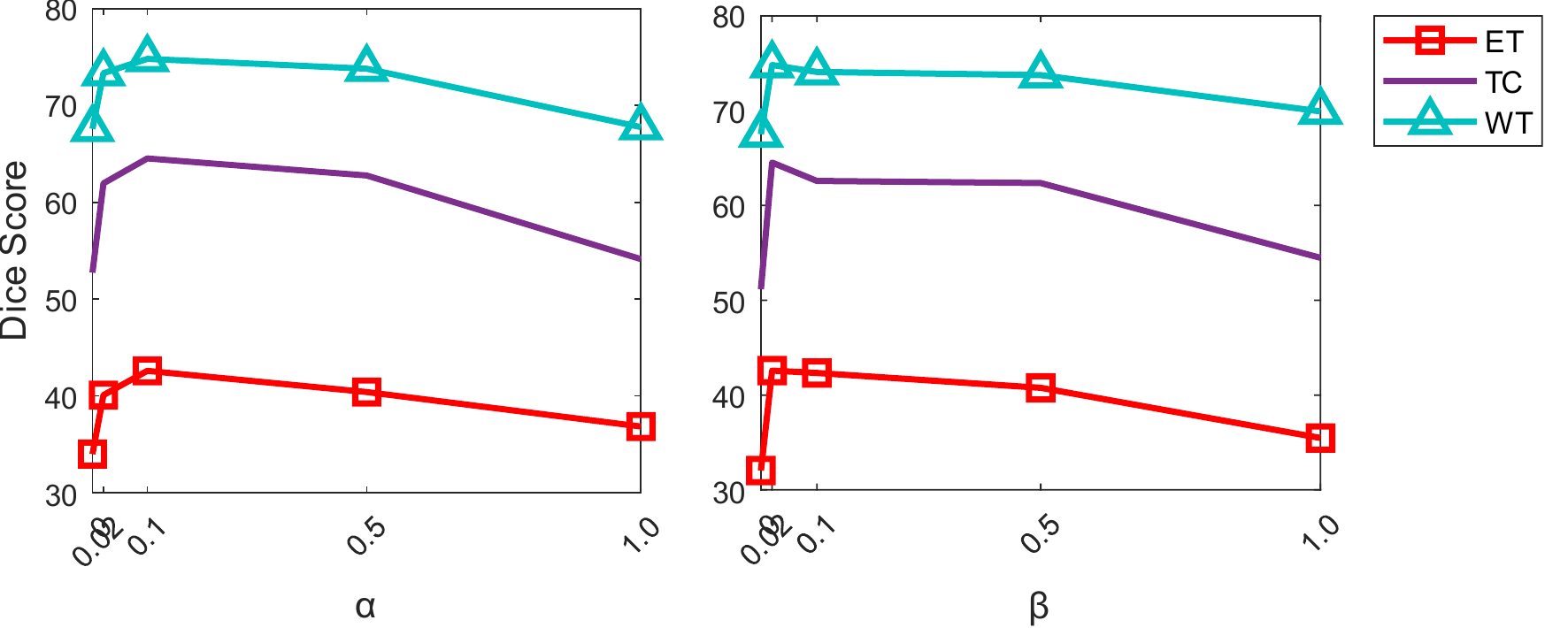}
\end{center}
\vspace{-5mm}
\caption{Sensitivity of $\alpha$ and $\beta$ for~\cref{eq:main_loss} for  non-dedicated training, where only T1 is available for evaluation on BraTS2018.}
\label{fig:sent-test}
\end{figure}

\noindent\textbf{Sensitivity of~\cref{eq:main_loss} to $\alpha$ and $\beta$:} The results are shown in~\cref{fig:sent-test} by setting the $\alpha$ and $\beta$ values to \{0,0.02,0.1,0.5,0.7,1\}. When testing $\alpha$ values, we set $\beta$ to 0.02; and when testing $\beta$, we set $\alpha$ to 0.1.
When $\alpha=\beta=1$, the results drop significantly, which can be explained by the scale of different loss functions, where too large weights for the auxiliary losses may disturb the gradient flow of the main task. 
In general, $\alpha=0.1$ and $\beta=0.02$ produces the best results.
Small values for the weights of the auxiliary tasks contribute to the whole process, but do not interfere with the main task optimisation. 
Interestingly, when $\alpha=0$ (only specific features are learned), the model can still segment the tumours to some extent by simple concatenation of specific features, which means that the specific features contain rich information. 
A similar conclusion can be reached when $\beta=0$ (only shared features are learned).

\noindent\textbf{Computational comparison:} We compare ShaSpec with SMIL (using the official code released by Ma et al.
) in terms of the number of model parameters, training/inference iteration time and GPU memory usage, where both models are trained/tested with batch-size of 4, and we estimate the average time consumption of 30 iterations on one 3090 GPU for a fair comparison.
SMIL has 0.33M parameters, with training iterations and testing taking 0.1309s and 0.0019s, and during training and testing, the GPU memory usage started from 1430MiB, climbed to 24268MiB, and then casted an ``out of memory'' error in the end.
On the other hand, the ShaSpec model has 0.22M parameters, takes 0.0257s for model training iteration and 0.0016s for model testing, and constantly consumes 1421MiB of GPU memory.

\noindent\textbf{An additional classification experiment on X-ray + clinical texts:} We conducted an extra classification experiment on OpenI~\cite{demner2016preparing}. We reorganised the OpenI dataset by only considering frontal images as visual inputs, and the `COMPARISON' and `FINDINGS' tags in the reports for textual inputs. Also, we reformulate the multi-label Chest-Xray classification into a binary classification problem (whether any chest problem exists or not). In total, the dataset contains 3851 pairs of visual-textual samples, which are split into 80\% for training and 20\% for evaluation. We adopt ResNet50 as visual backbone and an LSTM model with 128 hidden neurons as textual backbone.
The single-modal ResNet50 obtained AUC=0.77 with image inputs only; and the single-modal LSTM has AUC=0.86 with text inputs only. The baseline multi-modal model with simple feature concatenation trained/evaluated with full modality reaches AUC=0.90. When trained with 30\% missing image modality and evaluated only on texts, it achieves AUC=0.87. Our ShaSpec model shows better performance than the baseline model with AUC=0.89 (close to full modality).

\begin{figure}[t]
\begin{center}
\includegraphics[width=0.35\textwidth]{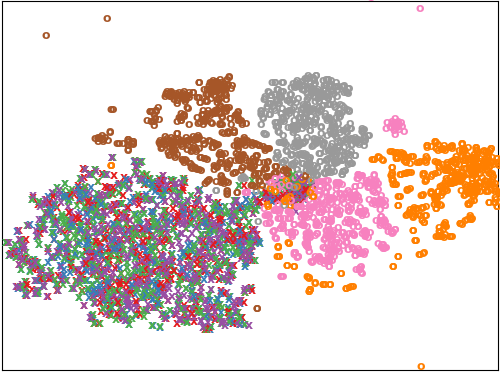}
\end{center}
\vspace{-3mm}
\caption{t-SNE visualisation of shared and specific features of four modalities from all training data on BraTS2018. The shared features of four modalities are presented by `\textbf{x}' in different colours, while the specific features of four modalities are presented by `\textbf{o}' in different colours.
}
\label{fig:vis-4modes}
\end{figure}

\noindent\textbf{Shared and Specific Feature Visualisation:} To further display the effectiveness of ShaSpec, we visualise the shared 
and specific features with t-SNE in~\cref{fig:vis-4modes}.
Note that the shared features 
are clustered together, while the specific features 
are well-separated, which shows that both types of features are exactly what is expected from our algorithm.

\section{Conclusion}

In this paper, we propose the simple but effective ShaSpec method to address multi-modal learning with missing modalities in training/testing, for dedicated/non-dedicated training, and applied to segmentation and classification tasks. 
Empirically, it outperforms the state-of-the-art by a large margin in different tasks and settings 
because of the semantically rich shared and specific learned features that are strongly related to the main task.
Through the t-SNE visualisation of the shared and specific feature space, we further verify the effectiveness of the method.
In the future, we will test the proposed ShaSpec method on other tasks (e.g., regression) and datasets to further verify its generality and effectiveness.

{\small
\bibliographystyle{ieee_fullname}
\bibliography{mybib}
}

\end{document}